\documentclass[conference]{IEEEtran}

\IEEEoverridecommandlockouts

\usepackage{cite}

\ifCLASSINFOpdf
  \usepackage[pdftex]{graphicx}
  \graphicspath{{./img/}}
  \DeclareGraphicsExtensions{.pdf,.jpeg,.png}
\else
\fi

\usepackage{amsmath}
\usepackage{amssymb}
\usepackage{mathtools}

\usepackage{url}

\usepackage{subcaption}
\usepackage[bottom]{footmisc}
\usepackage{hyperref}
\usepackage[noabbrev]{cleveref}
\usepackage{cuted}
\usepackage{xcolor}
\usepackage{booktabs}

\begin{document}

\title{Multi-Object Self-Supervised Depth Denoising}

\author{\IEEEauthorblockN{Claudius Kienle} 
	\IEEEauthorblockA{\textit{Karlsruhe Institute of Technology}\\
		Karlsruhe, Germany\\
	Email: claudius.kienle@student.kit.edu\\
                (privat@claudiuskienle.de)}
	\and
	\IEEEauthorblockN{David Petri}
	\IEEEauthorblockA{\textit{Karlsruhe Institute of Technology}\\
		Karlsruhe, Germany\\
Email: david.petri@student.kit.edu\\
(me@dapetri.com)}}

\maketitle

\begin{abstract}
	Depth cameras are frequently used in robotic manipulation, e.g. for visual servoing. The quality of small and compact depth cameras is though often not sufficient for depth reconstruction, which is required for precise tracking in and perception of the robot's working space. Based on the work of Shabanov et al. \cite{shabanov2020self}, in this work, we present a self-supervised multi-object depth denoising pipeline, that uses depth maps of higher-quality sensors as close-to-ground-truth supervisory signals to denoise depth maps coming from a lower-quality sensor. We display a computationally efficient way to align sets of two frame pairs in space and retrieve a frame-based multi-object mask, in order to receive a clean labeled dataset to train a denoising neural network on. The implementation of our presented work can be found at \href{https://github.com/alr-internship/self-supervised-depth-denoising}{https://github.com/alr-internship/self-supervised-depth-denoising}.
\end{abstract}

\IEEEpeerreviewmaketitle

\section{Introduction}

RGB-D sensors capture not only the RGB color information but also a depth value for each pixel. In the context of robotics, this bears the potential of better perception and manipulation in human environments. But just as the prices vary between RGB-D sensors, so does the quality of their depth maps. 

Economic sensors, such as the Intel RealSense product line, 
offer  compact optics capable of high frame rates which makes them more applicable for use on robot systems in dynamic environments. Yet consequently they lead to rather poor and noisy depth maps. More expensive sensors, such as those offered by Zivid, come with elaborate optics capable of capturing high-resolution depth maps subject to only very little noise. Yet consequently these sensors are bulkier, heavier, and due to their capturing technique only capable of a low frame rate. For these reasons, higher-quality sensors may be less feasible for the application on robots. In \Cref{tab:evaluation-comparison} the qualitative differences between such lower- and higher-quality RGB-D images and in the appendix in \Cref{fig:appendix:compare-cons} examples of such lower- and higher-quality sensors can be seen.\\

\begin{figure}[t]
    \includegraphics[width=\linewidth]{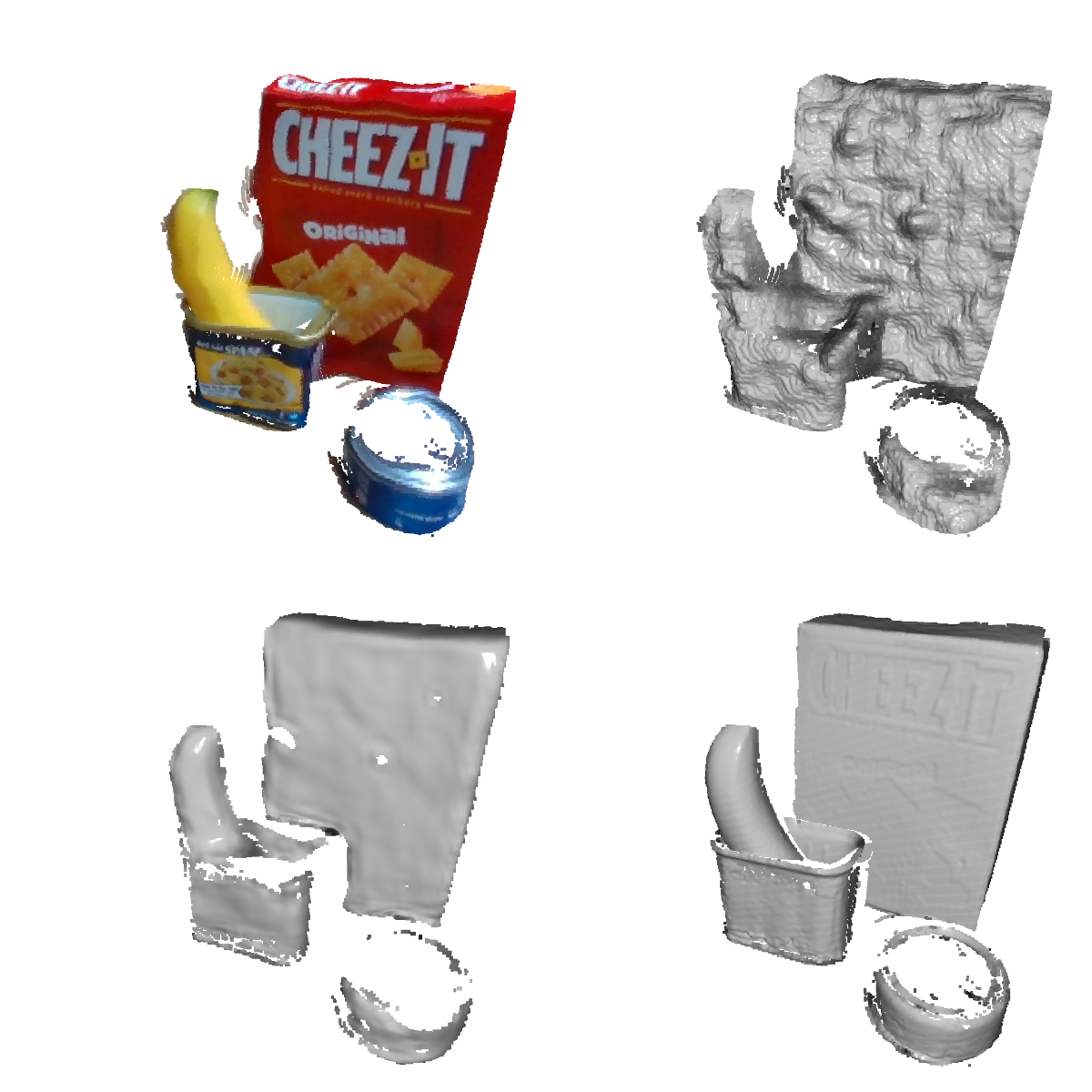}
    \caption{ This figure visualizes a sample inferred by the model trained on the augmented dataset. The upper row depicts the LQ RGB and depth frame. In the bottom row, the HQ depth frame is visualized on the right next to the predicted depth frame on the left. All frames are cropped to visualize the objects only. }
    \label{fig:eval-sample}
\end{figure}


In the context of this work, we build onto the self-supervised approach for the denoising of depth maps as presented by Shabanov et al. \cite{shabanov2020self}. After giving an introduction to related work in \Cref{sec:relatedwork}, the main body of this work is made up of two parts, namely dataset generation and neural network training. The process of dataset generation is explained in \Cref{sec:camset} where we chose YCB Objects \cite{calli2015benchmarking} as subjects for our multi-object scenes. We reduce the computational overhead introduced by the recalculation of the extrinsic transformation for every new sequence, which is explained in \Cref{sec:cal}. In order for our network to omit noisy background depth-values and to only focus on the objects, we applied a masking technique presented in \Cref{sec:mask}. As RGB-D sensors we use a RealSense D435 (RS) as lower-quality (LQ) sensor and a Zivid One+ (Zivid) as higher-quality (HQ) sensor. Due to Zivid's low frame rate, dynamic video capturing is not possible, which requires an efficient augmentation technique to increase the dataset size, which we elaborate in \Cref{sec:aug}. \Cref{sec:net} explains the implementation and training of the network used. Finally, we discuss our results in \Cref{sec:results} and compare them to those obtained by Shabanov et al. \cite{shabanov2020self}.

Overall, our work significantly differs from Shabanov et al. \cite{shabanov2020self} and contributes in the three following aspects: we present (1) a simple and straightforward approach for generating the near-to-ground-truth data in a computationally more efficient way from two sensors capturing the same scene. (2) Our approach does not rely on frame sequences but rather on a frame-to-frame basis, alleviating the need for high frame-rate cameras. And lastly (3) we present a augmentation technique in point-cloud space enabling us to effectively increase the training data-set while only relying on frame-to-frame image capturing.

\section{Related Work}
\label{sec:relatedwork}

Denoising depth maps captured by RGB-D cameras is a relatively new and small field of research. The denoising approaches can generally be divided
into two groups. On the one hand, classic non-machine-learning-based methods and, on the other hand, newer machine-learning-based methods. 

The classic approaches utilize hand-crafted filters for denoising. A rather old but very popular approach are Bilateral Filters (BF) introduced by Tomasi et al. \cite{tomasi1998bilateral} in 1998.
Bilateral Filters are non-linear filters that are designed to preserve high-contrast edges and remove low-contrast or gradual changes by relying on both spatial distance and photometric intensity difference between pixels. A more recent non-ml-approach are Rolling Guidance Filter (RGF) by Zhang et al. \cite{zhang2014} from 2014, which work similarly to Bilateral Filters but additionally take the scale of the filtered domain into account. 
These classic approaches do not need to be trained on any problem-specific dataset which makes their application straightforward for any problem domain. But thus by lacking any domain awareness these type of filters are unable to reconstruct and not only smoothen-out structural details typical to the domain.

Data-driven approaches are e.g. \textit{Depth Denoising and Refinement Network} by Yan et al. \cite{yan2018ddrnet} whose proposed framework tackles the denoising of depth maps by splitting the task into two subtasks. First aiming to denoise the low-frequency domain through self-supervised learning with a UNet-like architecture, using near-to-ground-truth depth maps, and second to refine the high-frequency domain through unsupervised learning by a shading-based criterion built on inverse rendering. \textit{Self-Supervised Deep Depth Denoising} (DDD) is a self-supervised approach introduced by Sterzentsenko et al. \cite{sterzentsenko2019self} based on a fully convolutional deep
Autoencoder with the intent to exploit the photometric and depth information of different points of view on the same scene, where view synthesis is used as
a supervisory signal. Yet another self-supervised approach is \textit{Self-Supervised Depth Denoising using Lower- and Higher-quality RGB-D sensors} (SSDD) by Shabanov et al. \cite{shabanov2020self} where the network for the denoising of lower-quality depth images is trained by using frames recorded by a higher-quality sensor of the same view as a close-to-ground-truth supervisory signal.
These data-driven approaches are capable of reconstructing domain-specific object details. This though comes at the cost of a complex and potentially domain and camera-specific data-preprocessing and training of the models.
 
\section{Methods} \label{sec:methods}

\begin{figure}[t] 
	\centering
	\includegraphics[width=0.4\textwidth]{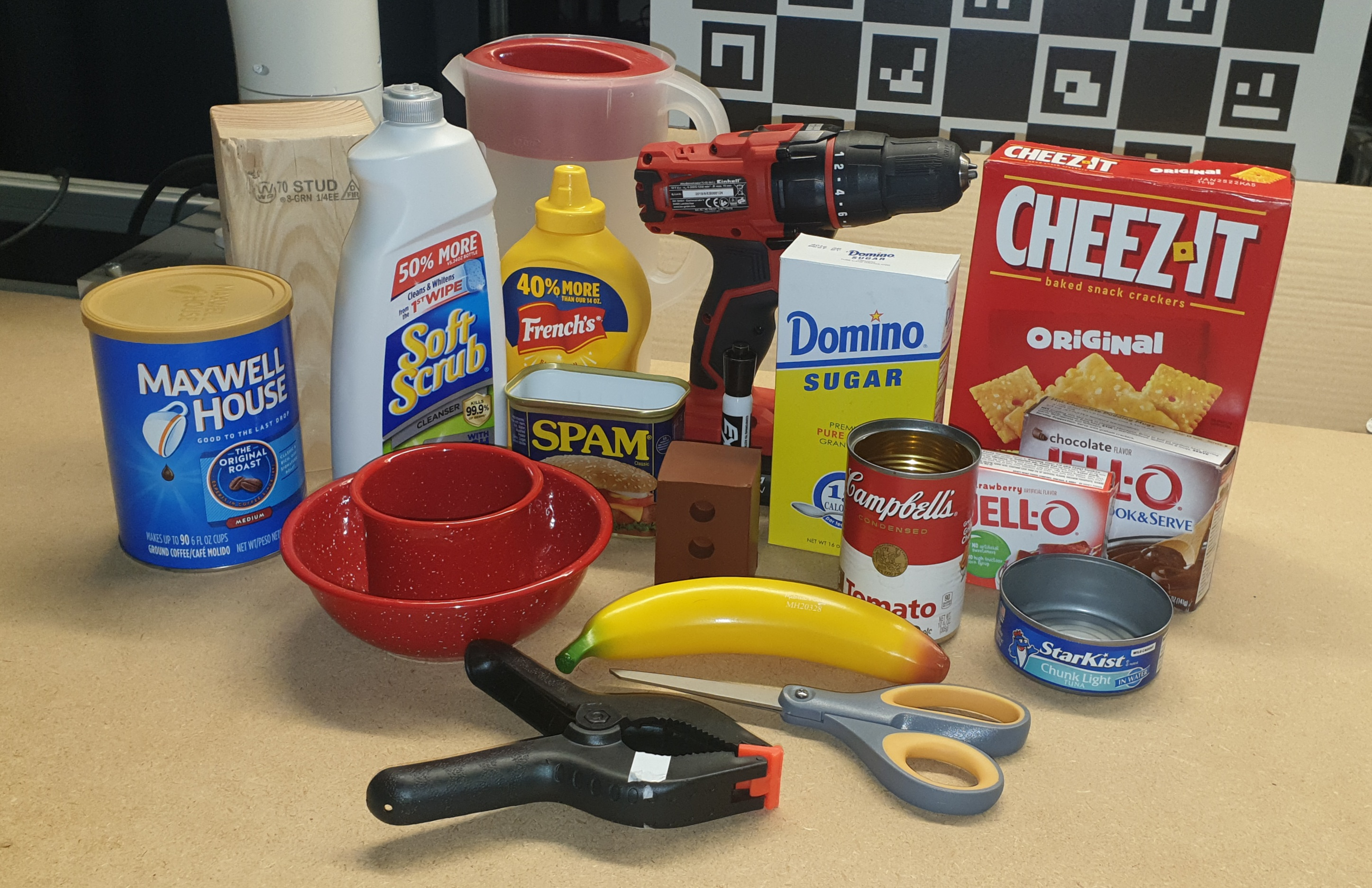}
	\caption[]{Image of the reduced set of 20 YCB objects used as objects of interest in this work.}
	\label{fig:ycb}
\end{figure}

The first part of the pipeline consists of the dataset generation, on which the networks will later be trained on. This first requires the setup of the sensors which we introduce in \cref{sec:camset}. \Cref{sec:cal} presents the calibration in use, which eliminates the naturally different view angles of the two cameras. After the cameras are calibrated correctly resulting in a pixel-wise correspondence of the RGB-D frames, we recorded the RGB-D dataset on a reduced set of 20 YCB Objects, shown in \cref{fig:ycb}.

\subsection{Dataset Acquisition} \label{sec:camset}

We recorded our dataset with a RealSense D435 (RS) as lower-quality (LQ) sensor and a Zivid One+ (Zivid) as higher-quality (HQ) sensor. 
Our acquisition setup can be seen in \Cref{fig:appendix:compare-cons}. The construction chosen should ensure that both sensors are mounted in the very same relative position across several mounts resulting in consistent data over several acquisitions. In addition, the height and placement of the pedestal were chosen to minimize the difference between the viewing angles of both cameras a priori, placing the sensors of both cameras on a common plane. This includes the angle position of the Zivid One+, as its viewing angle is shifted by 8.5 degrees counter-clockwise. 

For dataset acquisition, we captured all 20 YCB objects individually and also in composition with up to three additional objects. For each shot, we moved and rotated the objects randomly. This ensures a large enough dataset for network training. The raw dataset can be seen as a set of $N$ tuples:
$$\{(\boldsymbol{C}_{LQ}^i, \boldsymbol{D}_{LQ}^i, \boldsymbol{C}_{HQ}^i, \boldsymbol{D}_{HQ}^i):i=1,\cdots,N\}$$ where $\boldsymbol{C}^i$ contains the RGB color information and $\boldsymbol{D}^i$ the depth information of the $i$-th LQ and HQ images respectively. Our final dataset after acquisition contains $N=1024$ such tuples.

\subsection{Calibration} \label{sec:cal}

\begin{figure}[t]
	\centering
	\includegraphics[width=0.45\textwidth]{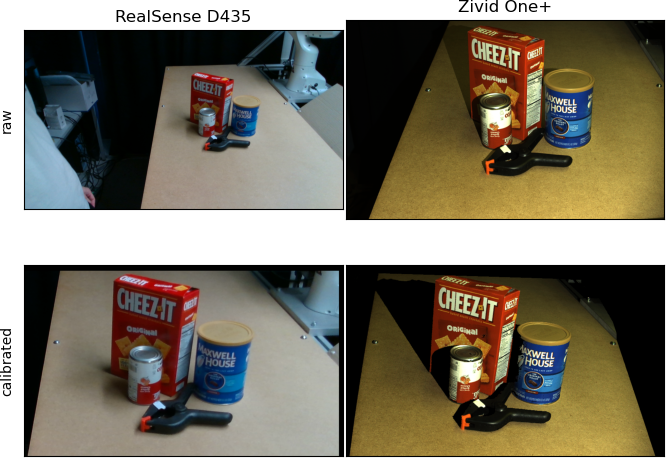}
	\caption{ The upper row displays the raw images of both cameras. The LQ frame is on the left and the HQ frame on the right. It is easy to see that the cameras have different fields of view. The respective calibrated images are visualized below. The rotation and translation that were applied during calibration align the objects but also increase the shadows of the objects in the Zivid frames.
		}
	\label{fig:raw-calibrated-images}
\end{figure}

Due to the fact that both sensors are mounted side by side, the resulting images show the captured objects from slightly different angles with hugely different fields of view, as shown in \cref{fig:raw-calibrated-images}. 
To facilitate the learning of the task at hand for the network, an extrinsic transformation must be applied. The extrinsic transformation equals a transform in 3D space that maps an HQ image onto the LQ sensor's plane, resulting in a pixel-wise correspondence of the frames. 

We are leveraging on the fact that our cameras can be mounted in the same relative position repeatedly, which reduces the calibration algorithm massively. Consequently, only a single transformation must be calculated, which avoids the associated computational overhead and also eliminates a possible source of error. Since the calibration is applied in 3D space, the images must be unprotected with the help of the respective depth maps, transformed, and finally reprojected into the image plane.

First, we unproject the color frame $\boldsymbol{C}$ with the depth frame $\boldsymbol{D}$ from image coordinates to camera coordinates. Formula \cref{eq:pinhole1,eq:pinhole2,eq:pinhole3} executes the unprojection.
\begin{align}
	x & = \frac{(u - c_x) * z}{f_x} 
	\label{eq:pinhole1}\\
	y & = \frac{(v - c_y) * z}{f_y} 
	\label{eq:pinhole2} \\
	z & = d * d_{scale}             
	\label{eq:pinhole3}
\end{align}
A pixel $p$ is defined by its row and column index $(u, v)$. The depth value $d$ of pixel $p$ is retrieved from the depth frame $\boldsymbol{D}$. Every point in the point cloud is colored with the RGB value at pixel $p$ given by the color image $\boldsymbol{C}$. Focal length $f_x$ and $f_y$, as well as the principal point $[c_x, c_y]^T$ are the camera's intrinsic parameters. This unprojection results in point clouds $\boldsymbol{P}$ for each LQ and HQ frame pair.
\begin{equation}
	(\boldsymbol{C}_{LQ},\boldsymbol{D}_{LQ}), (\boldsymbol{C}_{HQ},\boldsymbol{D}_{HQ})\xRightarrow[unproject]{} \boldsymbol{P}_{LQ}, \boldsymbol{P}_{HQ}
	\label{eq:2d_to_3d}
\end{equation}

The next step is to apply an extrinsic transformation on $\boldsymbol{P}_{HQ}$ so that it lines up with $\boldsymbol{P}_{LQ}$. We use a ChArUco board to select $K$ points $\{(\boldsymbol{p}_{HQ}^i, \boldsymbol{p}_{LQ}^i) \mid i = 1,\ldots,K\}$ that should match up after transformation. The optimal transformation matrix $\boldsymbol{\hat{T}}_{ex}$ is then given by the solution of the least-squares problem given in  \cref{eq:least-squares-points} \cite{arun1987least}. 
\begin{equation}
	\hat{\boldsymbol{T}}_{ex} = 
        \underset{\boldsymbol{T}}{\mathrm{arg\,min}} \sum_{k=1}^K 
            || \boldsymbol{T} * \left[\begin{array}{@{}c@{}}\boldsymbol{p}_{HQ}^i \\ 1\end{array}\right]
            	- \left[\begin{array}{@{}c@{}}\boldsymbol{p}_{LQ}^i \\ 1 \end{array}\right] ||
	\label{eq:least-squares-points}
\end{equation}
To refine the transformation, we attempted to use ICP \cite{rusinkiewicz2001efficient} to further align the point clouds, but due to the rather large dissimilarity between the point clouds, the refined transformation resulted in a worse alignment. 

The extrinsic transformation matrix $\boldsymbol{T}_{ex}$ can then be applied to $\boldsymbol{P}_{HQ}$.
\begin{equation}
	\boldsymbol{P}_{HQ} \xRightarrow[Tansf.]{extr.} \boldsymbol{\widetilde{P}}_{HQ}
	\label{eq:3d_transfrom}
\end{equation}
The transformed point cloud $\boldsymbol{\widetilde{P}}_{HQ}$ now matches the position and rotation of the point cloud $\boldsymbol{P}_{LQ}$.

Finally, $\boldsymbol{\widetilde{P}}_{HQ}$ can be reprojected back onto LQ's image plane, which is just the inverse operation of \cref{eq:pinhole1,eq:pinhole2,eq:pinhole3}, resulting in aligned color and depth frames. For this projection, only the intrinsics of the LQ camera are needed, since the HQ's point clouds are now transformed into the LQ's plane.
\begin{equation}
	\boldsymbol{\widetilde{P}}_{HQ} \xRightarrow[repoject]{} \boldsymbol{\widetilde{C}}_{HQ}, \boldsymbol{\widetilde{D}}_{HQ}
	\label{eq:3d_to_2d}
\end{equation}

\subsection{Masking}
\label{sec:mask}

\begin{figure*}[t]
    \centering
    \begin{subfigure}{0.245\textwidth}
        \includegraphics[width=\textwidth]{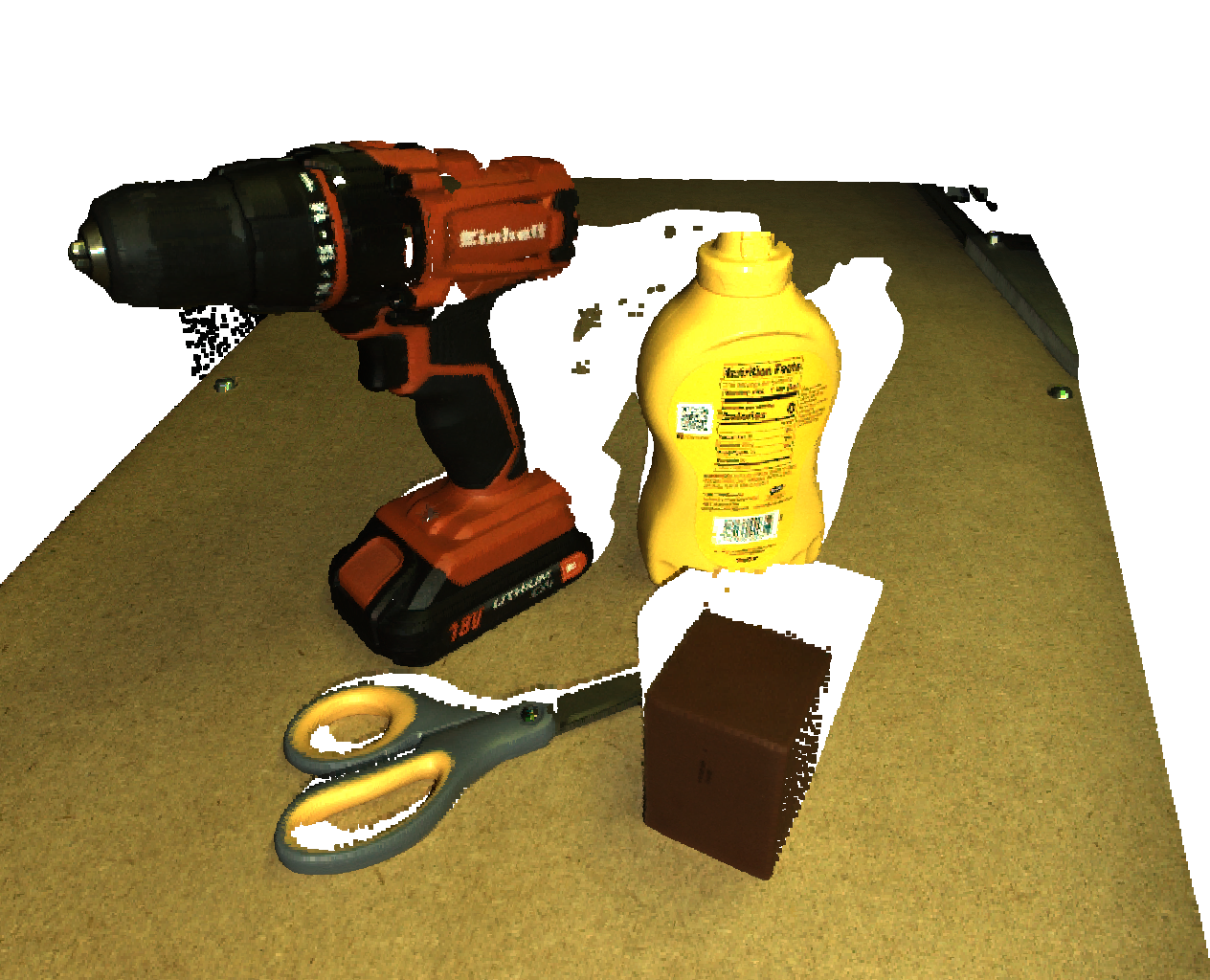}
        \caption{raw}
    \end{subfigure}
    \begin{subfigure}{0.245\textwidth}
        \includegraphics[width=\linewidth]{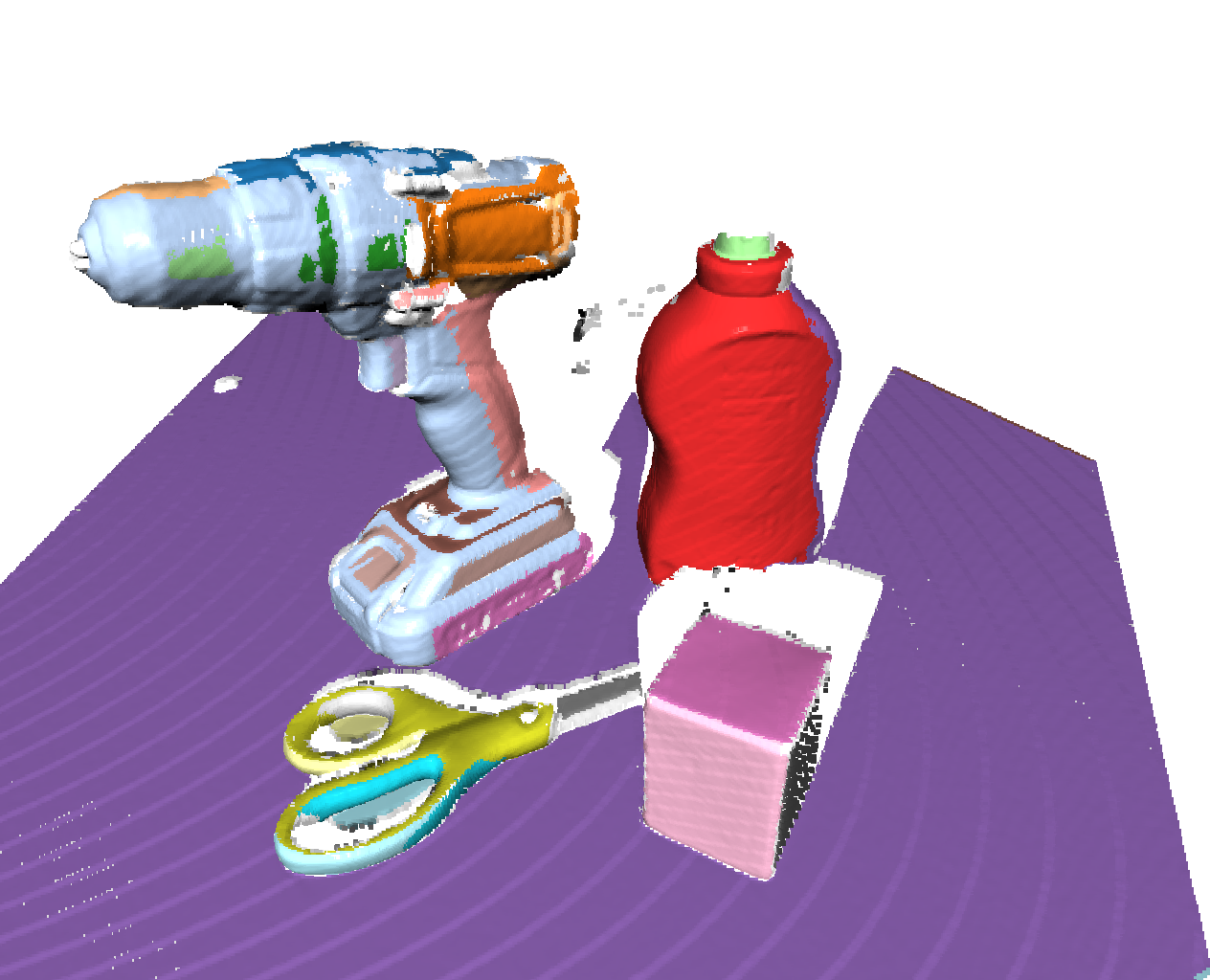}
        \label{fig:mask_nb_region_growing}
        \caption{NB}
    \end{subfigure}
    \begin{subfigure}{0.245\textwidth}
        \includegraphics[width=\linewidth]{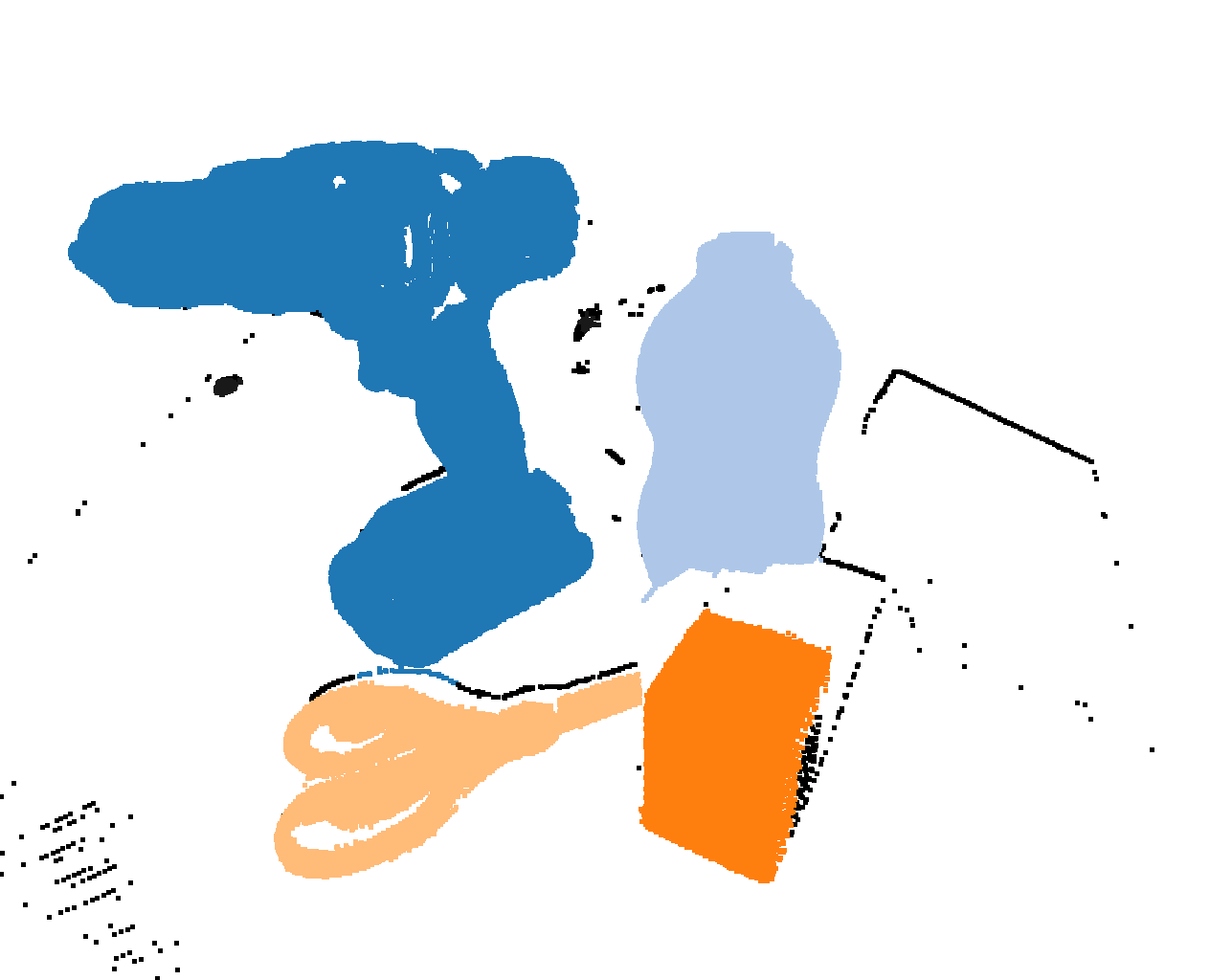}
        \label{fig:mask_db_region_growing}
        \caption{DB}
    \end{subfigure}
    \begin{subfigure}{0.245\textwidth}
        \includegraphics[width=\linewidth]{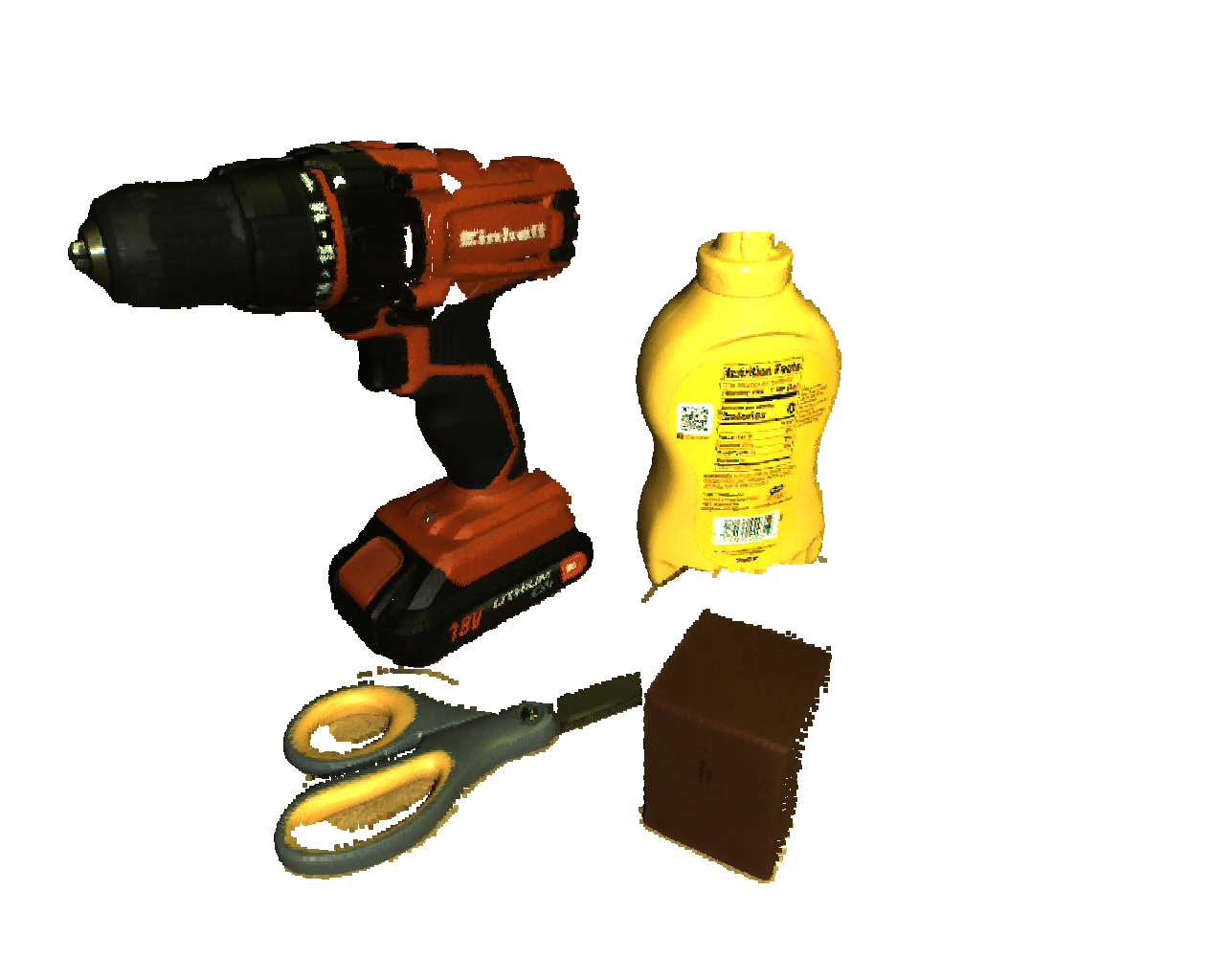}
        \caption{result}
    \end{subfigure}
    \caption{Masking Pipeline}
\end{figure*}

Since we are only interested in denoising the YCB objects, we compute a mask of these objects to later black out all background pixels. We implemented a classical algorithm to generate accurate masks for our dataset. The algorithm computes the mask with the help of the point clouds $\boldsymbol{P}_{HQ}$, $\boldsymbol{P}_{LQ}$, whereby the major part of the algorithm only works on the HQ point cloud. To increase performance, we first crop the HQ point cloud by a bounding box to a known region where the objects are located in. To separate the objects from the surface, a normal-based region growing is applied that clusters the point cloud into smooth surfaces as sown in \Cref{fig:mask_nb_region_growing}.
The region growing consequently operates on the normal difference of adjacent points. We estimate the normals of each point with open3d. Following, all clusters that surpass a threshold to a priori defined points get rejected. All remaining points are passed to the second stage of the masking pipeline.

Secondly, a density-based spatial clustering \cite{schubert2017dbscan} is applied, which aims to remove clusters that are too far away from the largest cluster or just too small to form a cluster on its own. Therefore, this clustering mainly removes outliers. A clustered point cloud can be seen in \Cref{fig:mask_db_region_growing}.
By reprojecting the HQ point cloud back into color and depth frames, the missing pixels form the new mask for the HQ frames. We additionally refine the mask with density-based clustering on the LQ point cloud, since it does not line up  perfectly with the HQ frame. This results in a second mask respective to the LQ frames. The resulting mask equals the intersection of both previous masks. The results of this masking algorithm can be seen in \Cref{fig:mask_db_region_growing}.
	
\subsection{Augmentation}
\label{sec:aug}
Due to the low frame rate of the Zivid camera to produce high-resolution depth images, dataset acquisition is comparably time-consuming. As a result, our final dataset consists only of 1024 frame pairs. We deployed heavy data augmentation to prevent our model from overfitting. 
Common data augmentation like random cropping would falsify our dataset since the depth values stored in the depth frames would not be augmented. For a rotation, translation, or scale, this results in depth values that do not match the position in the frame. For this reason, we used a different augmentation technique. Instead of augmenting the frames of each data tuple, we applied the augmentation in three-dimensional space. Each tuple $(\boldsymbol{C}_{LQ}, \boldsymbol{D}_{LQ}, \boldsymbol{C}_{HQ}, \boldsymbol{D}_{HQ})$ is unprojected to point clouds $(\boldsymbol{P}_{LQ}, \boldsymbol{P}_{HQ})$. Then $K$ randomly sampled transformations $\boldsymbol{T}_{rand}^i$ get applied on both of the point clouds. After reprojection of the newly generated point clouds, this augmentation results in a total of $K+1$ tuples for each data tuple. The augmentations were generated with a maximal translation of 10 cm and a maximal rotation of 5 degrees.

\subsection{Network \& Training} \label{sec:net}

Our raw dataset contains 1,024 image pairs and is significantly smaller than the dataset of Shabanov et al. \cite{shabanov2020self} with approximately 46,000 frame pairs. Additionally, the amount of temporal information held between consecutive frame pairs in our case is much smaller, if existent, due to not capturing dynamic image sequences. Splitting our small dataset into three equally sized sets $P_1, P_2, P_{test}$ for the OOF prediction scheme would result in a raw training set size of 340 for each model of the first level. That being a too small number to train any generalizable model, 
We trained a UNet network with 90 \% of our data corpus, splitting the remaining 10 \% equally in validation and test dataset, whereby the test dataset was only used for network evaluation.

Our denoising network is a UNet architecture, originally proposed by Ronneberger et al. \cite{ronneberger2015u}, with additional skip connections in the downward path. 32 initial feature maps worked best, which is reasonable due to our small dataset.

The input of the network consists of the four RGB-D channels of the input image concatenated with the object mask as the fifth channel. The network's output on the other hand is a single channel containing the predicted depth at each pixel. Since valid depth values must be greater or equal to zero, we also implemented an ReLU output activation that maps all negative depth values to zero.
	
	
In addition to using the RMSProp optimizer for training, we also deployed a learning rate scheduler. To speed up training, we enabled automatic mixed precision (amp) and used network gradient scaling.
	
We examined multiple options for data preprocessing to improve network convergence and prediction results. For one, the first three input channels encoding the RGB values are scaled linearly to $[0, 1]$. We also tested normalizing the input and target depth channel to zero mean and unit variance, which resulted in a less accurate prediction. To ease network training, the computed object mask was directly applied on the input depth and RGB channels, whereby all pixels outside the object mask get replaced with NaN values.

We trained our network with different loss functions, namely mean L1 Loss, mean L2 Loss, and Huber loss. 
\begin{align}
	\mathcal{L}_1(D_{pred},\tilde{D}_{HQ}) = \frac{\sum_{i,j} |d^{ij}|_1}{\sum_{i,j} m^{ij}}                   
	\label{eq:loss-l1}\\
	m^{ij} := m^{ij}_{nan} * m^{ij}_{obj} \quad d^{ij} := (d^{ij}_{pred}-\tilde{d}^{ij}_{HQ}) * m^{ij}\nonumber
\end{align}

To find the best set of hyperparameters, we deployed random hyperparameter tuning.

\section{Results}
\label{sec:results}

\subsection{Hyperparameter Tuning}
We evaluated all 100 models on the same test set containing 51 samples that the networks never have seen before. The evaluation consists of multiple metrics which provide information about how well the models denoise the input depth frames. In order to evaluate how much the network denoises the input depth frame with respect to the target depth frame, we computed every metric on the input/target pair as well as on the prediction/target pair. 

The first metric that we use for evaluation is the mean L1 Loss also depicted in \Cref{eq:loss-l1}. This metric provides information on how well the model denoises the overall input depth frame, independent of how large the pixel-wise depth differences between input and target were initially. To obtain more insight into which depth differences the model denoises and by how much, we also evaluated the model against three other metrics. Those three metrics are variants of the mean L1 Loss, whereby the loss is not computed over the whole depth pair, but only on the depths, whose depth differences are in the interval $[\delta_{min}, \delta_{max})$. The interval, as well as the depth differences, are given in millimeters. For evaluation, we used three such metrics with intervals $[0mm, 10mm)$, $[10mm, 20mm)$, and $[20mm, inf)$.
                                                                                                
It can be seen that all models depicted reduce the overall mean L1 Loss. The median of the leftmost box plot is reduced from 9.8 mm for the input/target pairs, to 8.4 mm for the prediction/target pairs. Therefore, the predicted depth frame of this model is on average more similar to the target depth frame than the input depth frame. When inspecting the composition of the L1 Loss in more detail, it can interestingly be noticed that all trained models focus to denoise depth values with a depth difference above 10 mm. The best model reduces the mean L1 Loss of input/target for a depth difference between 10 and 20 mm from nearly 14 mm to 9.2 mm in the median, which equals a denoising of roughly 35 percent. This implies that the networks mainly denoise relatively large deviations from the target depth. 
                                                                                                
On the other hand, the models do not succeed to denoise finer deviations, which can be seen in the increasing L1 Loss in the interval of $[0mm, 10mm]$. Even the metric of the best model increases from approximately 4.3 mm to 5.6 mm. After further evaluation of the other models with respect to this metric, none of the 100 trained models during hyperparameter tuning succeeded to denoise those small depth differences between input and target.
                                
The best model resulting from hyperparameter tuning uses skip connections, mean L1 Loss as loss function, ReLU as output activation, 32 initial channels, and scales the input images by 0.5.

\subsection{Refined Evaluation}
                                                                                                
\begin{figure*}[ht]
    \centering
    \includegraphics[width=0.8\linewidth]{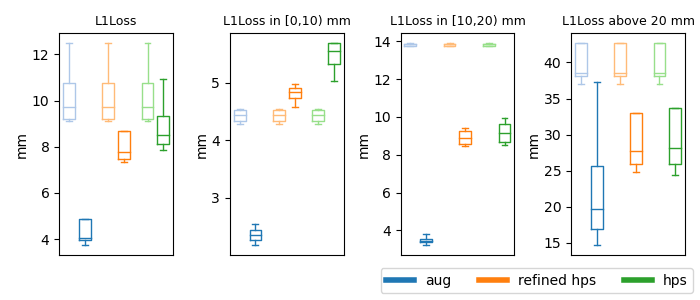}
    \caption{Model evaluation of three models trained with the same hyperparameters. The \textit{hps} plots display the evaluation of the best model obtained by hyperparameter search. We then refined this model by training it for another 24 epochs, whose evaluation is depicted as \textit{refined hps}. The leftmost plot \textit{aug} visualizes the evaluation of the model trained from scratch on the augmented dataset.}
    \label{fig:final-evaluation}
\end{figure*}
After determining the hyperparameters that resulted in the best evaluation, we trained two additional models with those parameters for a longer period, since the models for hyperparameter tuning were only trained for 25 epochs, which equals roughly one and a half hours. 
                                                                                                
We trained the two models on different datasets, one on the original dataset containing 1,024 samples and the other one on the augmented dataset containing 51,130 samples. The evaluation plots of both models are depicted in \Cref{fig:final-evaluation}. The plot also displays the best model from hyperparameter tuning for reference, which is named \textit{hps}.
The evaluation indicates that the model \textit{refine hps}, which was refined on the original dataset, denoises the depth difference above 10 mm not significantly better. However, the prediction/target mean L1 Loss for depth differences below 10 mm decreased significantly from 5.6 mm to 4.8 mm. This may be due to the longer training during which the learning rate was reduced more than while hyperparameter tuning.
                                                                                        
The evaluation of the model that was trained on the augmented dataset produces the best results.
We trained this model for 19 epochs, which took about 48 hours due to the large dataset. While this model trained on the augmented dataset, we evaluated it also on the non-augmented dataset to ease comparison and ensure consistent evaluation results. The median of all four prediction/target metrics for this model reduced notably in comparison to the other two models. Especially the L1 Loss for depth differences below 10 mm reduced from 4.4 mm to 2.4 mm, whereby the other models have not denoised this area at all. An input/target/prediction depth frame tuple is visualized in \Cref{fig:eval-sample}.
                                    
\begin{table*}[ht!]
    \centering
    \caption{Evaluation Comparison}
    \label{tab:evaluation-comparison}
    \begin{tabular}{l c c | c c c} \toprule & \multicolumn{2}{c}{Raw} & \multicolumn{3}{c}{Results}\\
                    & Shabanov & Ours   & Shabanov (basic) & Shabanov (LSTM) & Ours   \\
        \midrule                                                            
        MSE (mm)   & 57.22    & 261.17 & 31.61            & 21.02           & 103.74 \\
        IT/OT (\%) & -        & -      & 55.24            & 36.76           & 39.72  \\  \bottomrule
    \end{tabular}
\end{table*}
                                    
It can be seen that the predicted depth frame (bottom left) denoises the input depth frame. While the cereal box's surface in the LQ depth frame (top right) has irregular depth values compared to the HQ depth frame (bottom right), the neural network denoises the frame which results in a smoother surface. This can also be observed on other surfaces. Besides that, for regions where the observed scene contains jumps in the depth values, for example at the opening of the can, the input depth frame often means over these sharp edges while the target depth frame captures them more precisely. The predicted depth frame also started to learn those sharp jumps, as can be seen in \Cref{fig:eval-sample-top}. For this sample, the overall mean L1 Loss reduces from 6.03 mm for the input/target pair to 1.44 mm for the prediction/target pair.

\subsection{Comparison}
                                    
The comparison of our approach with the results from Shabanov et al. \cite{shabanov2020self} turns out to be very difficult for multiple reasons. 
                                    
On the one hand, we used a completely different dataset for network training and evaluation compared to them. While they trained the depth denoising of human bodies, we focused to denoise YCB Objects, which are considerably smaller, have more sharp edges, and are placed closer to the cameras. Our raw dataset, therefore, has an MSE of 261.17 mm compared to Shabanov et al. \cite{shabanov2020self} with an MSE of 57.22 mm. 
                                    
On the other hand, it is not clear if they computed the MSE on the whole depth frame or only on the mask. The former would result in a lower MSE, since the denominator also sums over pixels that got projected to 0 in the numerator. Shabanov et al. \cite{shabanov2020self} do not elaborate on how they compute the MSE for evaluation.
                                    
Therefore, in \Cref{tab:evaluation-comparison} we included a second metric named ``IT/OT'' besides the MSE in mm, which relates the MSE of each raw dataset with the respective results. It is computed by dividing the MSE of the results by the raw MSE and can be interpreted as how much noise is still present in the predicted depth frames. A value of 25 \% for example indicates that only 25 \% of the original noise is remaining.
                                    
For our approach, we evaluated the model trained on the augmented dataset since it performed best in comparison to the other models. We computed the MSE metric of our model again on the test set. While the prediction MSE of our approach, with a value of 103.74 mm, is worse than the MSE of Shabanov et al. \cite{shabanov2020self} with 21.02 mm, our raw dataset also has a larger MSE such that a comparison based on this metric is not sufficient. Considering the IT/OT metric, with a value of 39.72 \%, our approach is able to denoise the input image so that only 39.72 \% of the noise is left in the dataset. 

Compared to Shabanov et al. \cite{shabanov2020self}, our approach can be placed in the middle of their two approaches. Since we only used the first level of the two-level approach presented by them, our approach is more similar to the \textit{basic} model. Compared with this approach, our model reduces the noise significantly more with an IT/OT value of 36.76 \% compared to the 55.24 \% the evaluation of the \textit{basic} model achieved. The better performance of our model may be because the noise reduction between our input depth frames and the target depth frames is larger due to the higher quality camera we used.
 
\section{Conclusion}

We proposed a framework for data generation and self-supervised training of a network, with the goal of denoising depth frames originating from a lower-quality depth camera, using depth frames of a higher-quality depth camera as close-to-ground-truth data. Our approach is based on the framework proposed by Shabanov et al. \cite{shabanov2020self}, but is able to rely on fewer raw data frames, which additionally do not have to be timely correlated.
Our pipeline generates input/target RGB-D frame pairs of YCB Objects using simultaneously lower- and higher-quality sensors, which are then spatially aligned and included a mask of the to-be denoised YCB Objects. We applied a specially developed augmentation technique to increase the amount of training data, enabling us to rely on shot-by-shot data generation.

Our proposed framework, therefore, applies the work of Shabanov et al. \cite{shabanov2020self} for use in robotic manipulation and visual servoing. For this purpose, we adapted multiple steps of dataset generation and network training to our use case. To name a few, we simplified camera calibration by ensuring a fixed relative position, developed our own masking algorithm to reliably and precisely mask YCB Objects, and adapted the network architecture and training. 

\subsection{Discussion}

While our method is widely applicable, it also has some drawbacks.

To align the HQ frames to the LQ frames, a rotation in point cloud space is applied, which uncovers areas of the three-dimensional object, for which no pixel values in the two-dimensional image plane exist. These empty regions in the target frames make network training more difficult.

Another weak point of our pipeline is the mask generation. Our ability to generate such clear masks was thanks to the flat underground on which the YCB Objects were placed and the distant background in the image. These characteristics were exploited for our mask generation. The usage of high-quality masks was crucial for the generation of our results. The environment a robot observes in action most probably will not have such characteristics. Therefore, deploying our pipeline in action on a robot may result in poorer results and may require a revised masking tool.

Comparing our denoising task to the task of Shabanov et al. \cite{shabanov2020self}, we suggest that the denoising of YCB Objects is more difficult than of human bodies. YCB Objects, especially if multiple are closely grouped together, present many edges and smooth surfaces of different sizes. Human bodies on the other hand, especially as depicted in the paper of Shabanov et al. \cite{shabanov2020self}, present rather few edges and large surfaces of overall constant size. 

\subsection{Outlook}

It could be of interest to integrate our pipeline into an active robot system for permanent image capturing of the robot's working space. With a large enough dataset, the masking of the ROIs can then eventually be omitted, as the network could learn to denoise the whole input depth map. On the other hand, our masking algorithm could be applied to generate a masked YCB dataset without costly human segmentation. A segmentation model, which masks YCB Objects, could then be trained on this dataset. This might result in a more robust and real-time capable masking pipeline.

\bibliographystyle{IEEEtran}

\bibliography{IEEEabrv,bib/paper}
\clearpage

\appendix
\subsection{Aquisition Setup}

\begin{figure}[ht] 
	\centering
	\includegraphics[width=0.45\textwidth]{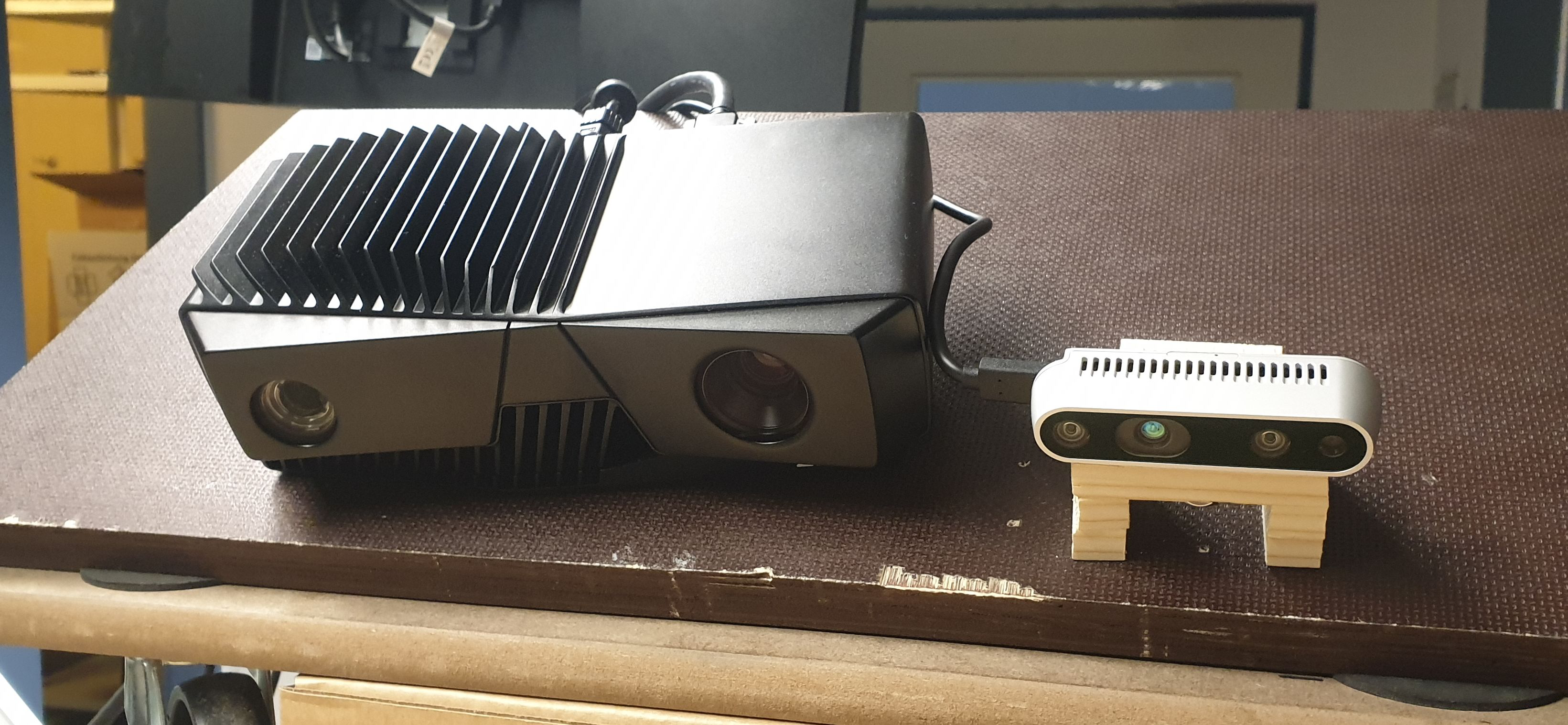}
	\caption[]{The picture depicts our acquisition setup. We mounted the two cameras next to each other on a wooden plate to prevent any relative positional changes during acquisition. Besides that, the cameras are positioned such that their field of view overlaps as much as possible.}
	\label{fig:appendix:compare-cons}
\end{figure}

\subsection{Additional Figures}

\begin{figure}[ht]
\centering
	\includegraphics[width=0.9\linewidth]{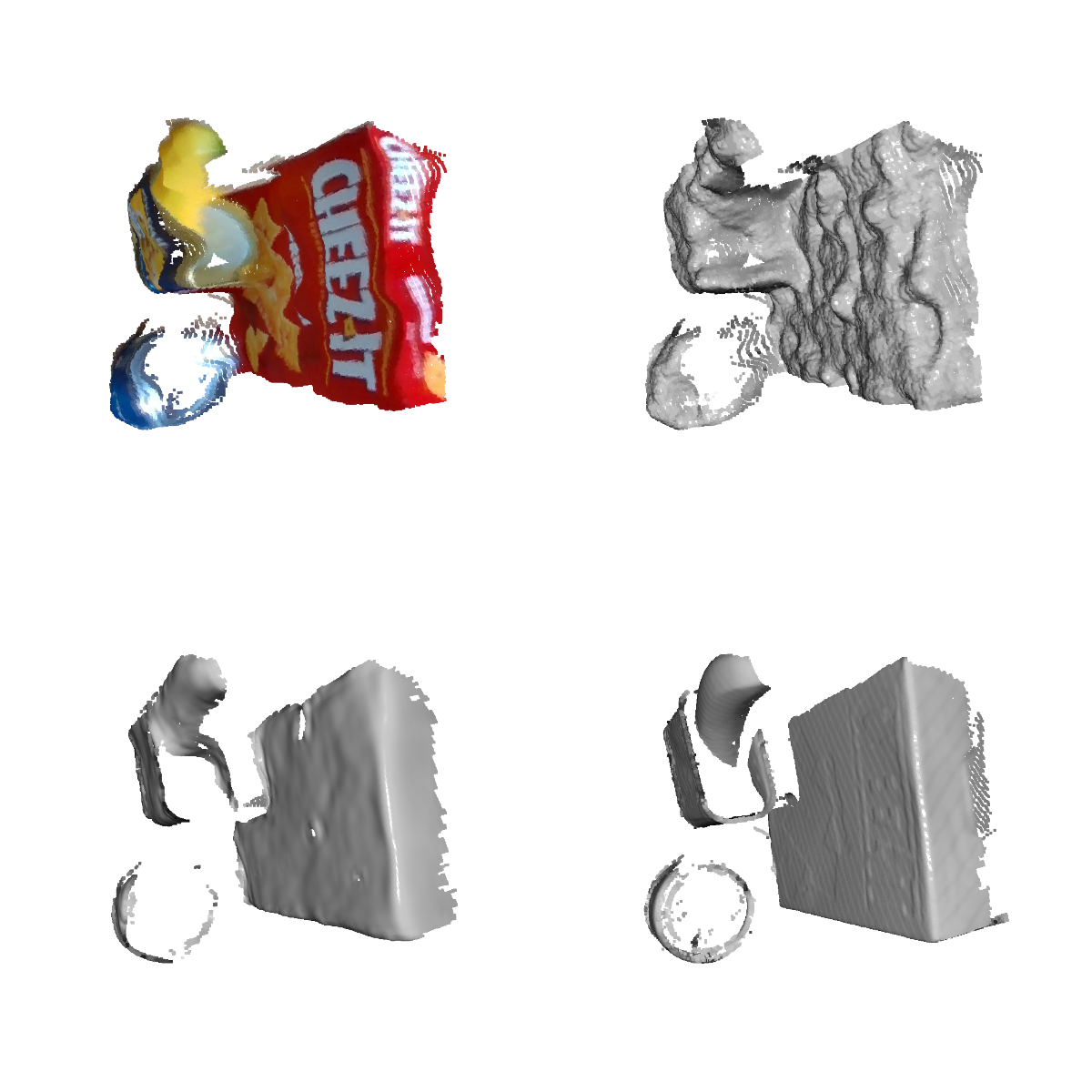}
	\caption{Top view of an input/prediction/target sample generated with the model trained on the augmented dataset. This visualization displays very clearly, how the network also learned to denoise sharp edges. While the input depth frame (top right) has many pixels at the opening of the can, the predicted depth frame reconstructs the sharp edges which can also be seen in the target depth frame.
		}
	\label{fig:eval-sample-top}
\end{figure}

\begin{figure} [h] 
\centering
	\includegraphics[width=0.8\linewidth]{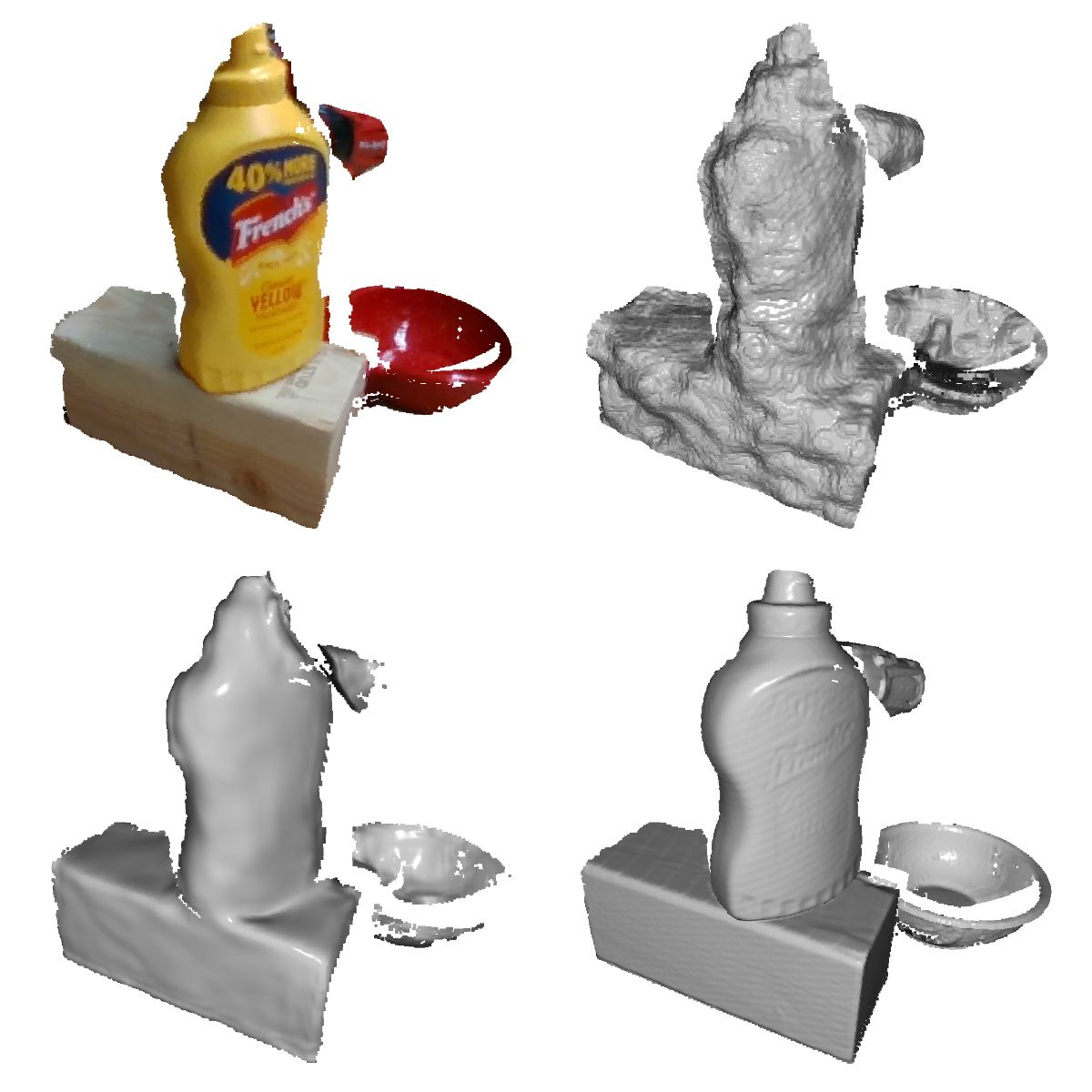}
 \caption*{------------------}
\end{figure}

\begin{figure}[h] 
\centering
	\includegraphics[width=0.8\linewidth]{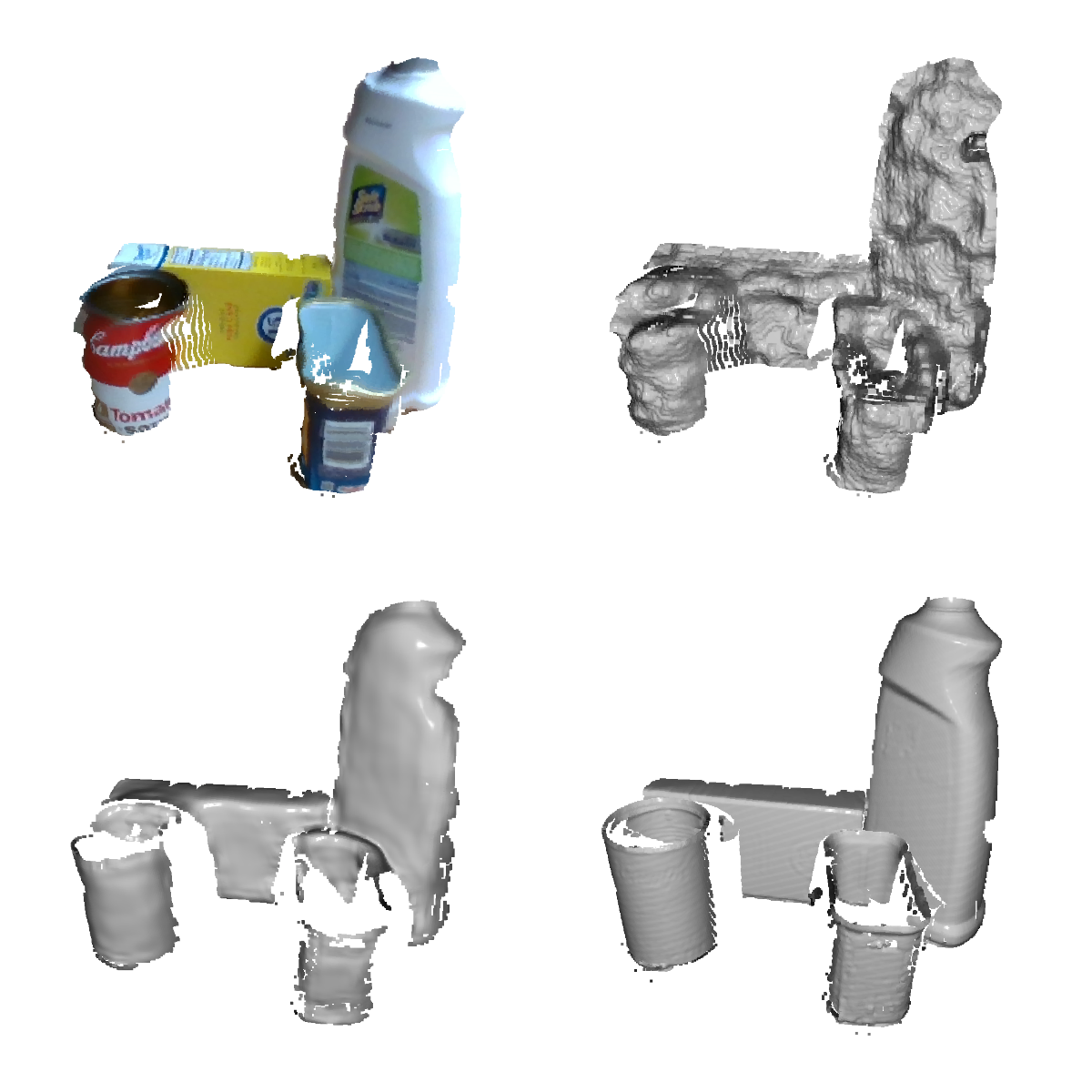}
	\caption*{------------------}
\end{figure}

\begin{figure}[h] 
\centering
	\includegraphics[width=0.8\linewidth]{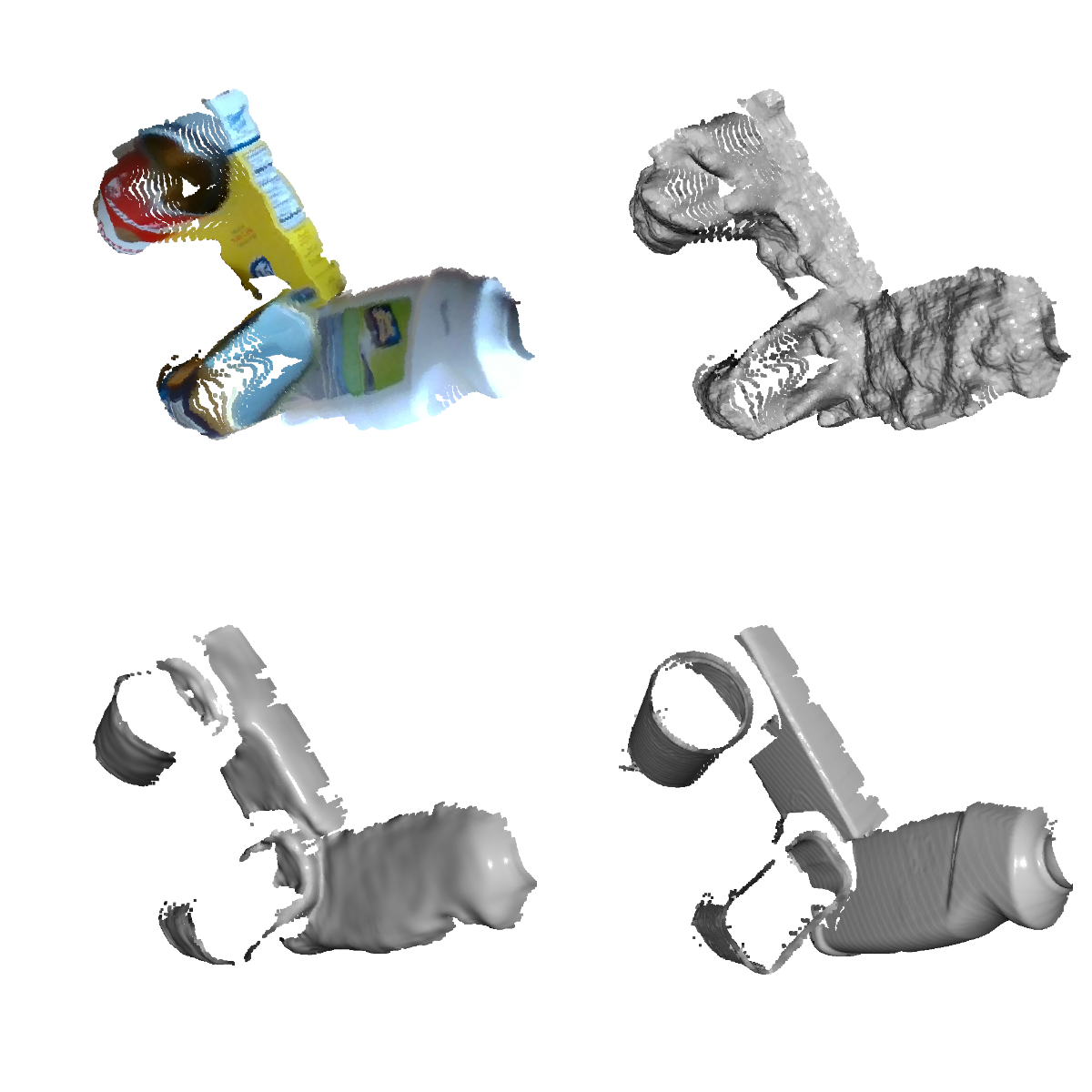}
 
\end{figure}

\end{document}